\documentclass[conference]{IEEEtran}
\IEEEoverridecommandlockouts
\usepackage{cite}
\usepackage{amsmath,amssymb,amsfonts}
\usepackage{algorithmic}
\usepackage{graphicx}
\usepackage{textcomp}
\usepackage{xcolor}
\usepackage{multirow}
\usepackage{hyperref} 
\usepackage{array,tabularx,makecell,booktabs}

\newcolumntype{Y}{>{\centering\arraybackslash}p{0.10\linewidth}} 
\def\BibTeX{{\rm B\kern-.05em{\sc i\kern-.025em b}\kern-.08em
    T\kern-.1667em\lower.7ex\hbox{E}\kern-.125emX}}
\begin{document}

\title{Exploration of Augmentation Strategies in Multi-modal Retrieval-Augmented Generation for the Biomedical Domain*\\
{\footnotesize \textsuperscript{*}A Case Study Evaluating Question Answering in Glycobiology}
\thanks{This work was supported by European Union under Horizon Europe [grant number 101159018] and EuroHPC JU [grant number 101101903]; by University of Maribor under the 2023 internal call Strengthening Researchers’ Programme Cores, field of Data Science and Artificial Intelligence in Biomedicine; Slovenian Research Agency [grant number GC-0001].}
}

\author{\IEEEauthorblockN{Primož Kocbek}
\IEEEauthorblockA{\textit{University of Maribor, Faculty of Health Sciences} \\
Maribor, Slovenia}
\IEEEauthorblockA{\textit{University of Ljubljana, Medical Factory} \\
Ljubljana, Slovenia \\
primoz.kocbek@um.si}
\and
\IEEEauthorblockN{Azra Frkatović-Hodžić}
\IEEEauthorblockA{\textit{Genos Ltd} \\
Zagreb, Croatia \\
afrkatovic@genos.hr}
\and
\IEEEauthorblockN{Dora Lalić}
\IEEEauthorblockA{\textit{Genos Ltd} \\
Zagreb, Croatia \\
dora@glycanage.com}
\and
\IEEEauthorblockN{Vivian Hui}
\IEEEauthorblockA{\textit{Center for Smart Health, School of Nursing} \\
\textit{The Hong Kong Polytechnic University}\\
Hong Kong, China \\
vivianc.hui@polyu.edu.hk }
\and
\IEEEauthorblockN{Gordan Lauc}
\IEEEauthorblockA{\textit{ University of Zagreb,} \\
\textit{Faculty of Pharmacy and Biochemistry} \\
Zagreb, Croatia }
\IEEEauthorblockA{\textit{Genos Ltd} \\
Zagreb, Croatia \\
glauc@genos.hr}
\and
\IEEEauthorblockN{Gregor Štiglic}
\IEEEauthorblockA{\textit{University of Maribor,} \\
\textit{Faculty of Health Sciences}\\
Maribor, Slovenia}
\IEEEauthorblockA{\textit{Usher Institute} \\
\textit{University of Edinburgh}\\
Edinburgh, UK \\
gregor.stiglic@um.si}
}

\maketitle

\begin{abstract}
Multi-modal retrieval-augmented generation (MM-RAG) promises grounded biomedical QA, but it is unclear when to (i) convert figures/tables into text versus (ii) use optical character recognition (OCR)-free visual retrieval that returns page images and leaves interpretation to the generator. We study this trade-off in glycobiology, a visually dense domain.
We built a benchmark of 120 multiple-choice questions (MCQs) from 25 papers, stratified by retrieval difficulty (easy text, medium figures/tables, hard cross-evidence). We implemented four augmentations—None, Text RAG, Multi-modal conversion, and late-interaction visual retrieval (ColPali)—using Docling parsing and Qdrant indexing. We evaluated mid-size open-source and frontier proprietary models (e.g., Gemma-3-27B-IT, GPT-4o family). Additional testing used the GPT-5 family and multiple visual retrievers (ColPali/ColQwen/ColFlor). Accuracy with Agresti–Coull 95\% confidence intervals (CIs) was computed over 5 runs per configuration.
With Gemma-3-27B-IT, Text and Multi-modal augmentation outperformed OCR-free retrieval (0.722–0.740 vs. 0.510 average accuracy). With GPT-4o, Multi-modal achieved 0.808, with Text 0.782 and ColPali 0.745 close behind; within-model differences were small. In follow-on experiments with the GPT-5 family, the best results with ColPali and ColFlor improved by ~2\% to 0.828 in both cases. In general across the GPT-5 family, ColPali, ColQwen, and ColFlor were statistically indistinguishable; ColFlor matched ColPali while being far smaller. GPT-5-nano trailed larger GPT-5 variants by roughly 8–10\%.
Pipeline choice is capacity-dependent: converting visuals to text lowers the reader burden and is more reliable for mid-size models, whereas OCR-free visual retrieval becomes competitive under frontier models. Among retrievers, ColFlor offers parity with heavier options at a smaller footprint, making it an efficient default when strong generators are available.
\end{abstract}

\IEEEpeerreviewmaketitle

\section{Introduction}

The rapid proliferation of Large Language Models (LLMs) has transformed numerous domains, including biomedical question answering (QA). Advanced systems such as Med-PaLM~2 have demonstrated expert-level performance on standardized medical examinations~\cite{singhal2023llms}, leveraging well-established biomedical benchmarks such as BioASQ~\cite{tsatsaronis2015bioasq} and PubMedQA~\cite{chen2019pubmedqa}. Despite these successes, their reliability in specialized scientific domains remains limited due to training data constraints. This highlights the need for external knowledge grounding to mitigate hallucinations and improve factual accuracy~\cite{ovadia2024finetuning}.

Retrieval-Augmented Generation (RAG) has emerged as a robust alternative or complement to fine-tuning (FT), dynamically providing relevant contextual information to LLMs during inference~\cite{lewis2020rag}. Compared with FT, RAG typically achieves stronger performance at substantially lower computational cost and does not require retraining when new knowledge becomes available~\cite{biomedrag2024}. This property makes RAG particularly suitable for rapidly evolving biomedical literature.

While text-based RAG systems are well established, scientific research often involves multi-modal content extending beyond text. Our study focuses on glycobiology—a visually dense and technically demanding domain encompassing complex molecular structures, pathway diagrams, and tabular datasets~\cite{frank2010glycobio}. Prior work has shown that even advanced LLMs struggle with glycobiology-related queries, frequently producing inconsistent or fabricated responses~\cite{williams2023glyco}. This underscores the need for multi-modal RAG (MM-RAG) systems capable of processing heterogeneous data modalities.

In this preliminary study, we compare two primary MM-RAG paradigms. The first converts multiple modalities into text following a conventional pipeline: PDF documents are parsed into structured elements—via direct text extraction or Optical Character Recognition (OCR)—while figures and tables are transformed into textual descriptions using summarization techniques~\cite{alkhalaf2024ehr}. The textual content is subsequently chunked and embedded for retrieval via standard vector databases. This approach simplifies decoupling between retrieval and generation but risks information loss and pipeline complexity.

The second paradigm employs vision-based document retrieval to circumvent conversion-related limitations. These methods treat entire document pages as images and utilize Vision-Language Models (vision-language models (VLMs)) to generate embeddings directly. A notable example is ColPali~\cite{faysse2025colpali}, which adapts the late-interaction mechanism from ColBERT~\cite{khattab2020colbert} to the visual domain, enabling fine-grained similarity matching between visual and textual embeddings. Related approaches such as VisRAG also leverage VLM-based retrievers and generators to preserve maximal information from original documents~\cite{yu2024visrag}. In such setups, the downstream LLM assumes greater responsibility for visual reasoning, effectively replacing traditional PDF parsing and summarization components.

This study investigates whether the performance of OCR-free, vision-based document retrieval depends on the multi-modal reasoning capacity and scale of the downstream generative model in the context of multiple-choice question (MCQ) answering from glycobiology literature. We evaluate three RAG strategies—standard text-based RAG, modality-converting MM-RAG, and vision-based ColPali—across both large proprietary models (e.g., the OpenAI GPT-4o family) and smaller open-source alternatives (e.g., Gemma-3-27B-IT). Furthermore, we assess different late-interaction visual retrievers—ColPali~\cite{faysse2025colpali}, ColFlor~\cite{masry2024colflor}, and ColQwen~\cite{colqwen2025}—using the recently released OpenAI GPT-5 family.

Our findings reveal a trade-off between pipeline simplicity and dependence on model reasoning capacity, establishing an initial baseline for developing reliable and trustworthy multi-modal RAG systems in specialized biomedical domains.

\section{Related Work}

The biomedical domain is inherently multi-modal, as clinicians and researchers routinely integrate information from medical imaging, laboratory results, electronic health records (EHRs), and genomics~\cite{google2023mmai, li2024multimodal}. This complexity makes it a natural setting for developing multi-modal artificial intelligence (AI) systems. Notable progress has been achieved in multi-modal question answering (MMQA), with models such as LLaVA-Med~\cite{li2023llavamed}, Med-Flamingo~\cite{moor2023medflamingo}, and Med-PaLM~M~\cite{tu2023generalist}, which jointly reason across textual and visual modalities in clinical contexts. However, public benchmarks have revealed potential data contamination in newer LLMs, motivating the use of private, domain-specific datasets for evaluation.

Biomedical AI represents a key application area for Retrieval-Augmented Generation (RAG), characterized by a vast and rapidly evolving knowledge base where factual accuracy and evidence-based reasoning are paramount~\cite{liu2025ragreview, aljunid2025ragreview}. Standard LLMs are static and prone to hallucinations, making RAG critical for building reliable clinical systems~\cite{liu2025ragreview}. A systematic review of 30 studies identified diagnostic support, EHR summarization, and medical QA as the most prevalent RAG applications~\cite{aljunid2025ragreview}. A meta-analysis of 20 studies comparing baseline LLMs with RAG-augmented counterparts reported a pooled odds ratio of 1.35 (95\% CI: 1.19–1.53), demonstrating significant performance gains through RAG integration~\cite{liu2025ragreview}.

For instance, a biomedical QA system employing a fine-tuned Mistral-7B model that retrieved information from PubMed and medical encyclopedias achieved a BERTScore F1 of 0.843~\cite{biomedragqa2025a, biomedragqa2025b}. Similarly, systems such as MEDGPT~\cite{sree2024medgpt} have demonstrated the practical utility of RAG for diagnostics and automated report generation.
Despite these advances, several challenges persist. The technical terminology and structural density of biomedical literature often introduce retrieval noise, resulting in suboptimal generation~\cite{aljunid2025ragreview}. Furthermore, biomedical content is fundamentally multi-modal: research articles, clinical guidelines, and EHRs frequently include visual components such as graphs, flowcharts, and tables that are integral to interpretation. Yet, most existing biomedical RAG implementations remain text-focused. The same systematic review confirming RAG’s clinical benefits revealed that only three of the twenty analyzed studies explicitly incorporated tabular or visual data—and typically by converting these modalities into text~\cite{liu2025ragreview}. This limited engagement with native multi-modal content underscores the need for architectures capable of directly processing complex visual information.

To address this gap, the MRAG-Bench benchmark was introduced to evaluate vision-centric multi-modal retrieval-augmented models~\cite{mragbench2025iclr}. It contains multiple-choice questions requiring visual reasoning across scenarios involving perspective shifts, occlusion, and temporal or geometric transformations~\cite{mragbench2025iclr}. The benchmark highlights a substantial gap between machine and human visual reasoning capabilities: when provided with ground-truth visual knowledge, GPT-4o’s accuracy improved by only 5.82\%, whereas human participants improved by 33.16\%~\cite{mragbench2025iclr}. These findings quantitatively support the hypothesis that the performance of vision-based RAG pipelines depends strongly on the scale and multi-modal reasoning ability of the underlying generative model.

\section{Materials and Methods}

\subsection{Benchmark}

A private benchmark dataset was constructed by two domain experts, comprising 120 multiple-choice questions (MCQs) derived from 25 original research and review manuscripts. Each question contained four possible answers and the corpus spanned core glycobiology concepts and applications, ranging from population-scale IgG glycomics, immune and inflammatory regulation, and endocrine/aging effects, to cardio-metabolic, gastrointestinal, pulmonary, and oncologic disease phenotypes (Appendix~\ref{app:dd}). Questions were also categorized by retrieval difficulty by consensus by the two domain experts: \emph{easy} when the answer appeared directly in the text, \emph{medium} when it was presented in tables or figures, and \emph{hard} when it required integrating information across text, figures, supplementary tables, or cited references. Iterative manual refinements were performed through review of model-generated explanations, during which five items were reclassified to ensure consistent difficulty labeling.

\subsection{Vision-Language Model Selection}

The selection of open-source vision-language models (VLMs) was constrained by local inference resources—specifically a single NVIDIA H100 80GB PCIe GPU. Considering activation memory and framework overhead, models up to approximately 30~billion parameters were feasible under 16-bit precision. We focused on models adapted from general multi-modal LLMs to biomedical applications~\cite{cheng2024domainmlm}, employing a generate-then-filter pipeline to synthesize diverse visual-instruction data from biomedical image–caption pairs. The evaluated models included Qwen2-VL-2B-Instruct~\cite{wang2024qwen2vl}, LLaVA-NeXT-Llama3-8B~\cite{li2024llavanelxt}, and Llama-3.2-11B-Vision-Instruct~\cite{meta2024llama32}. We additionally tested Google’s Gemma 3 model \texttt{gemma-3-27b-it}~\cite{kamath2025gemma3}. The were deployed using vLLM~\cite{kwon2023efficient} via docker (example in Appendix~\ref{app:bb})).

For proprietary baselines, the OpenAI GPT-4o family (\texttt{gpt-4o}, version~2024-11-20; \texttt{gpt-4o-mini}, version~2024-07-18) and GPT-5 family (\texttt{gpt-5}, \texttt{gpt-5-mini}, \texttt{gpt-5-nano}; all version~2025-08-07) were accessed via API under a GDPR-compliant Data Processing Addendum (DPA)~\cite{openai2025privacy}.

\subsection{Multi-Modal RAG Framework}

The multi-modal RAG framework consisted of three modular components: a document parser, an open-source vector store, and a vision-language model for text–visual alignment at inference. IBM’s \emph{Docling} served as the document parser~\cite{auer2024docling}, and \emph{Qdrant} as the vector database~\cite{qdrant2025}, supporting both single- and multi-vector representations. Four vector store configurations were developed: (i) \emph{text-only conversion}, in which all modalities were summarized into text; and (ii) \emph{visual late-interaction} variants (ColPali), in which document pages were represented as image embeddings. Retrieval employed a late-interaction mechanism~\cite{khattab2020colbert} for fine-grained similarity matching. Standard semantic embeddings (BAAI/bge-base-en-v1.5)~\cite{xiao2024cpack} were used for the text-only and multi-modal–text configurations. A general prompt template was used for creating table and figure summaries (Appendix~\ref{app:aa}).
Docling, Qdrant and the open-source models were deployed on-premise on a GPU server (NVIDIA H100 80GB PCIe GPU) using docker and accessed through api (example configurations are in Appendix~\ref{app:bb}). For chunking, we adopted Docling’s \texttt{HierarchicalChunker}~\cite{auer2024docling}, which uses the structural information encoded in the \texttt{DoclingDocument} to produce one chunk per detected document element. We set a token budget of 16{,}000 tokens and capped image resolution at 1{,}300 pixels on the longer side.

We compared multiple augmentation strategies supplied to the LLM: \emph{None} (query only), \emph{Text} (nearest text chunks via standard RAG), \emph{Multi-Modal} (raw figures and tables with corresponding summaries), and vision-based retrievals (\emph{ColPali}, \emph{ColQwen}, \emph{ColFlor}), in which the most similar document pages were retrieved and passed directly to the LLM.

\subsection{Vision-Based Retrievers}

We selected vision-based retrievers that demonstrated strong performance on the ViDoRe v2 benchmark~\cite{mace2025vidore}—particularly on healthcare-related subsets—and that were parameter-efficient, such as ColFlor. Specifically, we used the repositories \texttt{vidore/colpali-v1.3-merged} for ColPali, \texttt{vidore/colqwen2-v0.2} for ColQwen, and \texttt{ahmed-masry/ColFlor} for ColFlor.

\textbf{ColPali} (\texttt{vidore/colpali-v1.3-merged}) is a vision-based document retrieval model built on the PaliGemma-3B vision–language backbone. It extends SigLIP by feeding patch embeddings into PaliGemma to produce ColBERT-style multi-vector representations of pages~\cite{vidore_colpali_v1_3}. The model (2.92~billion parameters) encodes entire page images—including text, layout, figures, and tables—into patch embeddings, then matches queries to pages using late-interaction scoring (e.g., MaxSim) over token–patch pairs~\cite{hf_colpali_docs}. Strengths include preservation of visual layout and non-textual information, elimination of OCR and layout parsing, and strong retrieval accuracy on visual document benchmarks. Limitations include high memory and storage requirements due to multi-vector embeddings and increased computational cost for large document collections.

\textbf{ColQwen2} (\texttt{vidore/colqwen2-v0.2}) follows the ColPali paradigm but leverages the Qwen2-VL-2B backbone. It adopts a multi-vector retrieval scheme and incorporates adapters on top of Qwen2-VL-2B~\cite{manu_colqwen2_v0_2}. The Qwen2 backbone provides vision–language alignment across image and text modalities. Owing to its modular, adapter-based architecture, ColQwen2 facilitates efficient model switching and reduced update cost. However, its performance depends on the quality of adapter tuning; suboptimal alignment can degrade retrieval quality relative to ColPali. It also inherits the same storage and matching complexity constraints associated with multi-vector retrieval.

\textbf{ColFlor} (\texttt{ahmed-masry/ColFlor}) is a lightweight visual retriever optimized for footprint and speed. According to its model card, ColFlor contains approximately 174~million parameters—roughly 17$\times$ smaller than ColPali~\cite{hf_colflor_card}. It achieves query encoding $\sim$9.8$\times$ faster and image encoding $\sim$5.25$\times$ faster than ColPali, with only a $\sim$1.8\% performance drop on text-rich English documents~\cite{hf_colflor_blog}. Architecturally, ColFlor employs Florence-2’s DaViT vision encoder followed by a BART text encoder to produce contextualized visual embeddings, projected into a 128-dimensional latent space for retrieval via late interaction~\cite{hf_colflor_blog}. Advantages include low computational requirements, high throughput, and suitability for resource-constrained environments, whereas limitations include reduced performance on highly visual or non-English documents and limited capacity for capturing fine-grained visual detail due to its smaller backbone.

\subsection{Evaluation}

Performance was quantified using mean accuracy, defined as the proportion of correctly answered MCQs. With four equally likely options, random guessing yields an expected accuracy of 0.25. We evaluated both overall and difficulty-stratified performance. Two primary experiments were conducted with a general prompt template used (Appendix~\ref{app:cc}). 

First, we compared proprietary GPT-4o family of models with the open-source Gemma-3-27B-IT across all augmentation strategies, performing ten runs per configuration—five with permuted answer orders and five without—to assess potential benchmark contamination. The selection of open source model was due to the superior performance among other open-source models (Table~\ref{tab:augmentation_comparison}). Second, we compared vision-based retrievers (ColPali, ColQwen, ColFlor) using the GPT-5 family of models, with five evaluation runs each. In both cases we used default temperature and seed values - details for first case in Table~\ref{tab:augmentation_comparison}, which was gathered from model documentation and for the second case we note that the temperature parameter was removed from the GPT-5 family of models, seed was random. All accuracy  point estimates were reported using Agresti–Coull~\cite{agresti1998approximate} 95\% confidence intervals (CIs) aggregated across runs. This method provides a robust and stable estimation for binomial proportions, particularly when sample sizes are finite or proportions are near the interval boundaries of 0 or 1.

We checked for benchmark contamination by comparing answer-order permutation or shuffling. For each model-augmentation combination a paired t-test was performed with a conservative Bonferroni correction for multiple testing (Figure~\ref{fig:1}).

Because each augmentation–model combination was evaluated on the identical set of 120 MCQs, per-run accuracies form natural paired observations; we therefore used paired non-parametric tests (Wilcoxon signed-rank) for within-model comparisons with Bonferroni correction was used. The resulting test values and p-values are summarized in Table~\ref{tab:evaluations1_pvalues}.

We lastly looked at some retrieval metrics (precision@5), cost (cost per run, cost-per-correct comparison) and latency (tokens per second) analysis aggregated across runs focusing on vision-based retrievers and GPT-5 family of models. We reported point estimates using bootstrapped 95\% CIs aggregated across runs. 

\section{Results}

\begin{table}[htbp]
\centering
\caption{Initial model evaluation with regard to the proposed framework.}
\label{tab:augmentation_comparison}
\renewcommand{\arraystretch}{1.2}
\setlength{\tabcolsep}{1pt} 
\begin{tabularx}{\linewidth}{@{}X Y Y Y Y Y Y@{}}
\toprule
\textbf{Model} & \textbf{Temp.*} & \textbf{Seed*} &
\multicolumn{4}{c}{\textbf{Augmentation}} \\
\cmidrule(l){4-7}
& & & \textbf{None} & \textbf{Text} &
\makecell{\textbf{Multi}\\\textbf{-modal}} & \textbf{ColPali\textsuperscript{**}} \\
\midrule
\makecell[l]{AdaptLLM/biomed\\-LLaVA-NeXT-Llama3-8B}
  & 0.6 & 0   & 0.258 & 0.192 & 0.283 & 0.258 \\
\makecell[l]{AdaptLLM/biomed\\-Qwen2-VL-2B-Instruct}
  & 0.01 & 0  & 0.217 & 0.283 & 0.283 & /     \\
\makecell[l]{AdaptLLM/biomed\\-Llama-3.2-11B\\-Vision-Instruct}
  & 0.6 & 0   & 0.383 & 0.508 & 0.542 & 0.475 \\
\underline{google/gemma-3-27b-it}
  & 0.7 & 0   & \underline{0.483} & \underline{0.692} & \underline{0.767} & \underline{0.558} \\
gpt-4o-mini
  & 0.7 & rnd. & 0.458 & 0.708 & 0.675 & 0.558 \\
\textbf{gpt-4o}
  & 0.7 & rnd. & \textbf{0.592} & \textbf{0.792} & \textbf{0.833} & \textbf{0.758} \\
\bottomrule
\multicolumn{7}{p{\linewidth}}{\footnotesize
*default values of temperature and seeds were used (rnd. is random)  **AdaptLLM/biomed-Qwen2-VL-2B-Instruct ColPali augmentation did not produce relevant answers.}
\end{tabularx}
\end{table}

From the initial testing (Table~\ref{tab:augmentation_comparison}), smaller open-source models—even when fine-tuned—struggled on domain-specific MCQs. Augmentation improved accuracy by roughly 12–26\%. Among open-source models, \texttt{gemma-3-27b-it} achieved the strongest performance and was selected for detailed evaluation. ColPali augmentation performed well primarily with \texttt{gpt-4o}.

\begin{figure}[h]
    \centering
    \includegraphics[width=\linewidth]{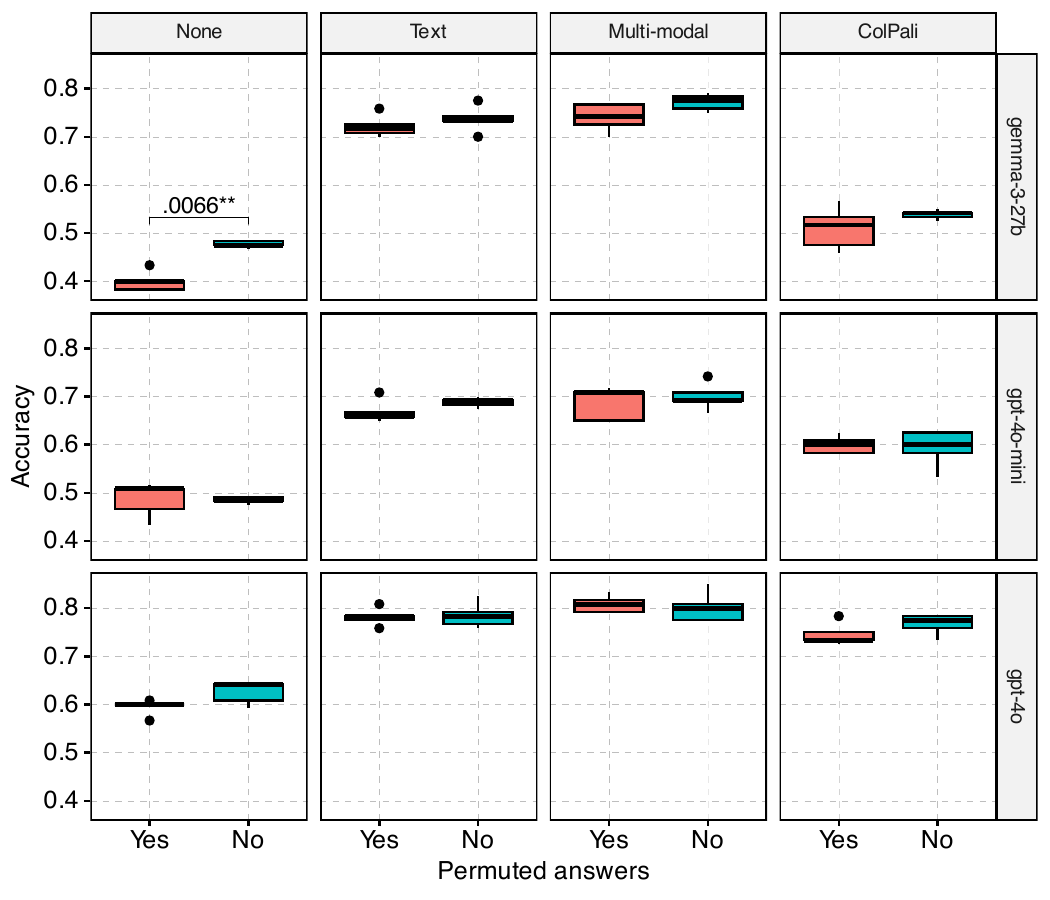}
    \caption{Boxplot and significant $p$-values for accuracy across selected models and augmentations.}
    \label{fig:1}
\end{figure}

Because contamination in public benchmarks can undermine evaluation robustness, we tested whether answer-order permutations affected performance. As expected (Figure~\ref{fig:1}), we found no statistically significant differences at $\alpha=0.05$. The closest case was \texttt{gemma-3-27b-it} with no augmentation ($p=0.0066$), consistent with its lower accuracy and higher variability (0.477; 95\% CI [0.437, 0.517]) versus 0.400 (95\% CI [0.363, 0.440]).

\hspace*{-6em}
\begin{table}[htbp]
\centering
\caption{Model evaluation of models with regard to the proposed framework.}
\label{tab:evaluations_1}
\renewcommand{\arraystretch}{1.2}
\setlength\tabcolsep{1 pt}
\begin{tabular}{llrrrr}
\hline
\textbf{Model} & \textbf{Aug.} &\begin{tabular}{c}\textbf{Easy}\\ \textbf{(n=69)}\end{tabular} & \begin{tabular}{c}\textbf{Medium}\\ \textbf{(n=24)} \end{tabular}& \begin{tabular}{c}\textbf{Hard}\\ \textbf{(n=27)} \end{tabular}&\begin{tabular}{c}\textbf{Average}\\ \textbf{(n=120)}\end{tabular}\\
\hline
\multirow{4}{*}{\begin{tabular}{c}\textbf{gemma-3}\\ \textbf{-27b-it}\end{tabular}} 
& None & 0.414  & 0.407 & 0.350  & 0.400  \\
&  & \scriptsize{[0.364, 0.467]} & \scriptsize{[0.328, 0.492]} & \scriptsize{[0.270, 0.439]} & \scriptsize{[0.362, 0.440]} \\
& Text & 0.780  & 0.711  & 0.567  & 0.722  \\
& & \scriptsize{[0.733, 0.820]} & \scriptsize{[0.629, 0.781]} & \scriptsize{[0.477, 0.652]} & \scriptsize{[0.684, 0.756]} \\
& Multi& \underline{0.786}  & \underline{0.741}  & \underline{0.608}  & \underline{0.740}  \\
& -modal & \scriptsize{\underline{[0.739, 0.826]}} & \scriptsize{\underline{[0.661, 0.808]}} & \scriptsize{\underline{[0.519, 0.691]}} & \scriptsize{\underline{[0.703, 0.774]}} \\
& ColPali & 0.548 & 0.459  & 0.458  & 0.510 \\
& & \scriptsize{[0.495, 0.600]} & \scriptsize{[0.377, 0.543]} & \scriptsize{[0.372, 0.547]} & \scriptsize{[0.470, 0.550]} \\
\hline
\multirow{3}{*}{\begin{tabular}{c}\textbf{gpt-4o}\\ \textbf{-mini}\end{tabular}} 
& None & 0.568 & 0.422  & 0.325 & 0.487 \\
&  &\scriptsize{[0.515, 0.619]} &  \scriptsize{[0.342, 0.507]} & \scriptsize{[0.248, 0.413]} & \scriptsize{[0.447, 0.527]} \\
& Text & 0.745 & 0.600  & 0.525  & 0.668  \\
&  & \scriptsize{[0.696, 0.788]} &  \scriptsize{[0.516, 0.679]} & \scriptsize{[0.436, 0.612]} & \scriptsize{[0.630, 0.705]} \\
& Multi & \underline{0.754}  & 0.563 & \underline{0.633} & \underline{0.687}  \\
& -modal &  \scriptsize{\underline{[0.705, 0.796]}} &\scriptsize{[0.479, 0.644]} & \scriptsize{\underline{[0.544, 0.714]}} & \scriptsize{\underline{[0.648, 0.723]}} \\
& ColPali & 0.609 & \underline{0.667} & 0.500 & 0.600 \\
&  &\scriptsize{[0.556, 0.659]} & \scriptsize{\underline{[0.583, 0.741]} }&\scriptsize{[0.412, 0.588]} & \scriptsize{[0.560, 0.638]} \\
\hline
\multirow{4}{*}{\textbf{gpt-4o}} 
& None & 0.664  & 0.630  & 0.358 & 0.595  \\
&  & \scriptsize{[0.612, 0.712]} & \scriptsize{[0.546, 0.707]} &  \scriptsize{[0.278, 0.447]} & \scriptsize{[0.555, 0.634]} \\
& Text & 0.832 & 0.830 & 0.583  & 0.782  \\
&  & \scriptsize{[0.789, 0.868]} &  \scriptsize{[0.757, 0.884]} & \scriptsize{[0.494, 0.668]} &  \scriptsize{[0.747, 0.813]} \\
& Multi & \textbf{0.835} & \textbf{0.874} & \textbf{0.658} & \textbf{0.808} \\
& -modal & \scriptsize{\textbf{ [0.792, 0.870]}} & \scriptsize{\textbf{ [0.807, 0.921]}} & \scriptsize{\textbf{ [0.570, 0.737]}} & \scriptsize{\textbf{ [0.775, 0.838]}} \\
& ColPali & 0.748  & 0.837  & 0.633  & 0.745 \\
& & \scriptsize{[0.699, 0.791]} & \scriptsize{[0.765, 0.891]} & \scriptsize{[0.544, 0.714]} & \scriptsize{[0.709, 0.778]} \\
\hline
\end{tabular}
{\parbox{3.3in}{
\vspace{0.2cm}
\footnotesize Note: [95\% CI] - Agresti-Coull 95\% confidence interval; underline - best performing augmentation for open-source model(gemma3-27b-it); bold - best performing augmentation for proprietary model (gpt-4o); Aug. - augmentation.}
}
\end{table}

In the first main experiment, \texttt{gpt-4o} achieved the highest overall accuracy with multi-modal augmentation (0.808; 95\% CI [0.775, 0.838]) (Table~\ref{tab:evaluations_1}). Text augmentation (0.782; 95\% CI [0.747, 0.813]) and ColPali (0.745; 95\% CI [0.709, 0.778]) were slightly lower; pairwise differences were not statistically significant for \texttt{gpt-4o}. For \texttt{gemma-3-27b-it}, multi-modal (0.740; 95\% CI [0.703, 0.774]) and text (0.722; 95\% CI [0.684, 0.756]) significantly outperformed ColPali (0.510; 95\% CI [0.470, 0.550]). By difficulty, \texttt{gpt-4o} with ColPali (0.745; 95\% CI [0.709, 0.778]) was comparable to \texttt{gemma-3-27b-it} with its best augmentation (0.740; 95\% CI [0.703, 0.774]).

\begin{table}[htbp]
\centering
\caption{Pairwise comparisons of statistically significant difference between augmentation and models ($p<0.05$).}
\label{tab:evaluations1_pvalues}
\renewcommand{\arraystretch}{1.1}
\begin{tabular}{llcr}
\hline
\textbf{Model} & \textbf{Augmentations} & \textbf{Test-value (V)} &\textbf{p-value} \\
\hline
\multirow{5}{*}{\textbf{gemma-3-27b-it}} 
& None $<$ Text &597.0 & $<0.001$ \\
& None $<$ ColPali & 291.5& $<0.001$ \\
& None $<$ Multi-modal& 1078.5& $<0.001$ \\
& ColPali $<$ Text & 3448.5&$<0.001$ \\
& ColPali $<$ Multi-modal&3442.0 & $<0.001$ \\
\hline
\multirow{5}{*}{\textbf{gpt-4o-mini}} 
& None $<$ Text & 676.5&$<0.001$ \\
& None $<$ Multi-modal&597.0 & $<0.001$ \\
& None $<$ ColPali &718.0 &$<0.001$ \\
& ColPali $<$ Text & 1997.0&$0.012$ \\
& ColPali $<$ Multi-modal&2291.5 & $0.001$ \\
\hline
\multirow{4}{*}{\textbf{gpt-4o}} 
& None $<$ Text &319.0& $<0.001$ \\
& None $<$ Multi-modal&215.0 & $<0.001$ \\
& None $<$ ColPali &359.0 &$<0.001$ \\
& ColPali $<$ Multi-modal&959.5 & $0.014$ \\

\hline
\end{tabular}
\end{table}

Within-model model pairwise comparisons using the Wilcoxon signed-rank test with Bonferroni correction are summed in Table~\ref{tab:evaluations1_pvalues}. We can observe that all augmentations (Text, ColPali, Multi-modal) outperform no augmentation and only for \texttt{gpt-4o} Text augmentation does not outperformed ColPali.

\begin{table}[htbp]
\centering
\caption{Performance of visual retrievers (ColPali, ColQwen, ColFlor) across the GPT-5 model family. Values are mean accuracy with Agresti–Coull 95\% CIs.}
\label{tab:gpt5_retrievers}
\renewcommand{\arraystretch}{1.2}
\setlength\tabcolsep{0.8 pt}
\begin{tabular}{llrrrr}
\hline
\begin{tabular}{c}\textbf{LLM}\\ \textbf{model}\end{tabular}& \begin{tabular}{c}\textbf{Retrieval}\\ \textbf{model}\end{tabular}& \begin{tabular}{c}\textbf{Easy}\\ \textbf{(n=69)}\end{tabular} & \begin{tabular}{c}\textbf{Medium}\\ \textbf{(n=24)} \end{tabular}& \begin{tabular}{c}\textbf{Hard}\\ \textbf{(n=27)} \end{tabular}&\begin{tabular}{c}\textbf{Average}\\ \textbf{(n=120)}\end{tabular}\\
\hline
\multirow{3}{*}{\textbf{gpt-5}} 
& ColPali & \underline{0.835} & 0.830 & 0.808 & \textbf{0.828} \\
& & \underline{[0.792, 0.870]} & [0.757, 0.884] & [0.728, 0.869] & \textbf{[0.796, 0.856]} \\
& ColQwen & \underline{0.835} & 0.822 & 0.783 & 0.822 \\
& & \underline{[0.792, 0.870]} & [0.748, 0.878] &[0.701, 0.848] & [0.789, 0.850] \\
& ColFlor & 0.823 & 0.837 & \underline{0.833} & \textbf{0.828} \\
& & [0.779, 0.860] & [0.765, 0.891] & \underline{[0.756, 0.890]} & \textbf{[0.796, 0.856]} \\
\hline
\multirow{3}{*}{\begin{tabular}{c}\textbf{gpt-5}\\ \textbf{-mini}\end{tabular}} 
& ColPali & 0.817 & \underline{0.859} & 0.717 & 0.807 \\
& & [0.773, 0.855] & \underline{[0.790, 0.909]} & [0.630, 0.790] & [0.773, 0.836] \\
& ColQwen & 0.817 & 0.830 & 0.708 & 0.798 \\
& & [0.773, 0.855]&[0.757, 0.884] & [0.621, 0.782] & [0.764, 0.829] \\
& ColFlor & 0.823 & \underline{0.859} & 0.758 & \underline{0.818} \\
& & [0.779, 0.860] & \underline{[0.790, 0.909]} & [0.674, 0.827] & \underline{[0.785, 0.847]} \\
\hline
\multirow{3}{*}{\begin{tabular}{c}\textbf{gpt-5}\\ \textbf{-nano}\end{tabular}} 
& ColPali & 0.797 & 0.763 & 0.600 & \underline{0.750} \\
& & [0.751, 0.836] & [0.684, 0.827] & [0.511, 0.683] & \underline{[0.714, 0.783]} \\
& ColQwen & 0.786 & 0.793 & 0.575 & 0.745 \\
& & [0.739, 0.826] & [0.716, 0.853] & [0.486, 0.660] & [0.709, 0.778] \\
& ColFlor & 0.759 & 0.770 & 0.592 & 0.728 \\
& & [0.712, 0.802] & [0.692, 0.834] & [0.502, 0.675] & [0.691, 0.762] \\
\hline
\end{tabular}
{\parbox{3.3in}{
\vspace{0.2cm}
\footnotesize Note: [95\% CI] - Agresti-Coull 95\% confidence interval; underline - best performing retrieval model for a specific model; bold - best performing retrieval model and model combination}
}
\end{table}

In the second experiment (Table~\ref{tab:gpt5_retrievers}) comparing visual retrievers with the \texttt{gpt-5} family, the highest overall mean accuracy was tied between ColPali (0.828; 95\% CI [0.796, 0.856]) and ColFlor (0.828; 95\% CI [0.796, 0.856]) using \texttt{gpt-5}. For \texttt{gpt-5-mini}, medium-difficulty items reached the highest category-wise accuracy with ColPali and ColFlor (both 0.859; 95\% CI [0.790, 0.909]). In contrast, \texttt{gpt-5-nano} underperformed the larger variants by $\sim$8–10 \%.

Within each \texttt{gpt-5} model, differences among visual retrievers were small and not statistically significant. The smallest unadjusted $p$-value was 0.430 (ColPali vs.\ ColFlor on \texttt{gpt-5-nano}); after Bonferroni correction across within-model retriever comparisons, $p=1.0$.

\hspace*{-6em}
\begin{table}[htbp]
\centering
\caption{Performance of retrieval for the GPT-5 family with respect to latency and cost.}
\label{tab:retrieval_gpt5}
\renewcommand{\arraystretch}{1.2}
\setlength\tabcolsep{0.8 pt}
\begin{tabular}{llrrrrrr}
\hline
\footnotesize{\textbf{Model}} & \footnotesize{\begin{tabular}{c}\textbf{Retrieval}\\ \textbf{model}\end{tabular}}
& \footnotesize{\textbf{P@5} }
& \footnotesize{\textbf{Latency}}
& \footnotesize{\textbf{Tokens} }
& \footnotesize{\textbf{TTFT*}} 
& \footnotesize{\textbf{Cost**} }
& \footnotesize{\textbf{P/C}***} \\
\hline
\multirow{3}{*}{\textbf{gpt-5}} 
& \footnotesize{ColPali}
    & 0.020  & 20.12  & 4643.05  & 417.88  & 5.57  & 5.72 \\
& &\tiny{[0.000, 0.094]} 
  &\tiny{[0.00, 163.66]} 
  &\tiny{[4269.59, 5016.51]} 
  &\tiny{[0.00, 3739.00]} 
  &\tiny{[5.12, 6.02]} 
  &\tiny{[4.14, 7.29]} \\
& \footnotesize{ColQwen}
    & 0.025  & 22.76  & 4385.31  & 366.17  & 5.26  & 5.28 \\
& &\tiny{[0.017, 0.033]} 
  &\tiny{[0.00, 81.50]} 
  &\tiny{[4273.68, 4496.94]} 
  &\tiny{[0.00, 960.92]} 
  &\tiny{[5.13, 5.39]} 
  &\tiny{[5.08, 5.48]} \\
& \footnotesize{ColFlor}
    & 0.018  & 17.82  & 4626.25  & 431.14  & 5.55  & 5.67 \\
& &\tiny{[0.018, 0.018]} 
  &\tiny{[0.00, 151.84]} 
  &\tiny{[4378.85, 4873.66]} 
  &\tiny{[0.00, 3754.74]} 
  &\tiny{[5.25, 5.85]} 
  &\tiny{[5.36, 5.97]} \\
\hline
\multirow{3}{*}{\begin{tabular}{c}\textbf{gpt-5}\\ \textbf{-mini}\end{tabular}} 
& \footnotesize{ColPali}
    & 0.022  & 13.80  & 7787.21  & 962.55  & 1.87  & 1.87 \\
& &\tiny{[0.017, 0.026]} 
  &\tiny{[0.00, 44.56]} 
  &\tiny{[7112.34, 8462.07]} 
  &\tiny{[0.00, 2396.78]} 
  &\tiny{[1.71, 2.03]} 
  &\tiny{[1.59, 2.15]} \\
& \footnotesize{ColQwen}
    & 0.024  & 24.34  & 8920.67  & 754.34  & 2.14  & 2.23 \\
& &\tiny{[0.019, 0.030]} 
  &\tiny{[0.00, 89.54]} 
  &\tiny{[8867.94, 8973.40]} 
  &\tiny{[0.00, 2050.25]} 
  &\tiny{[2.13, 2.15]} 
  &\tiny{[1.98, 2.47]} \\
& \footnotesize{ColFlor}
    & 0.018  & 21.33  & 7592.96  & 388.19  & 1.82  & 1.89 \\
& &\tiny{[0.018, 0.018]} 
  &\tiny{[0.00, 85.57]} 
  &\tiny{[7562.46, 7623.45]} 
  &\tiny{[0.00, 1423.52]} 
  &\tiny{[1.82, 1.83]} 
  &\tiny{[1.65, 2.14]} \\
\hline
\multirow{3}{*}{\begin{tabular}{c}\textbf{gpt-5}\\ \textbf{-nano}\end{tabular}} 
& \footnotesize{ColPali}
    & 0.022  & 14.08  & 9909.67  & 1283.69  & 0.47  & 0.54 \\
& &\tiny{[0.011, 0.034]} 
  &\tiny{[0.00, 48.16]} 
  &\tiny{[9408.38, 10410.96]} 
  &\tiny{[0.00, 3391.69]} 
  &\tiny{[0.45, 0.50]} 
  &\tiny{[0.48, 0.59]} \\
& \footnotesize{ColQwen}
    & 0.026  & 23.66  & 11755.52 & 866.97  & 0.56  & 0.67 \\
& &\tiny{[0.024, 0.028]} 
  &\tiny{[0.00, 78.55]} 
  &\tiny{[11688.70, 11822.33]} 
  &\tiny{[0.77, 2364.71]} 
  &\tiny{[0.56, 0.57]} 
  &\tiny{[0.62, 0.72]} \\
& \footnotesize{ColFlor}
    & 0.018  & 13.07  & 10031.55 & 1363.26 & 0.48  & 0.55 \\
& &\tiny{[0.018, 0.018]} 
  &\tiny{[0.00, 44.49]} 
  &\tiny{[9928.94, 10134.17]} 
  &\tiny{[0.00, 3523.51]} 
  &\tiny{[0.48, 0.49]} 
  &\tiny{[0.48, 0.62]} \\
\hline
\end{tabular}

{\parbox{4.2in}{
\vspace{0.2cm}
\footnotesize
Note: [95\% CI] – bootstrap 95\% confidence intervals for each metric;\\ 
P@5 – precision at 5; *Throughput; ** Cost estimate for a run in USD;\\ ***Price-per-cost (US cents) – cost estimate per correctly answered question.}}
\end{table}

Because each question was derived from a single source article rather than a multi-document corpus, the study was not designed to primarily evaluate retrieval performance. Accordingly, retrieval quality, as measured by P@5 (Precision@5), was uniformly low with only modest variation across retrieval models (range 0.02–0.026; Table~\ref{tab:retrieval_gpt5}). This is expected for a single-document retrieval setting, as any additional returned documents beyond the relevant one are counted as false positives, which compresses P@5 even for strong visual retrievers.

In contrast, computational footprint and cost varied markedly across base models. Latency was similar across the GPT-5 family (13–24 s per request), but tokens per run increased as model size decreased, from roughly 4{,}500 for \texttt{gpt-5} to 8{,}000 for \texttt{gpt-5-mini} and about 11{,}000 for \texttt{gpt-5-nano}. Throughput followed the same trend with roughly 400, 700, and 1{,}300 tokens/s, respectively). Despite using fewer tokens, \texttt{gpt-5} was far more expensive, with mean per-run costs of \$5.6 versus \$1.9 and \$0.5, i.e., about 2.5× and 10× higher, respectively. Price-per-correctly answered query showed the same pattern: roughly 5.7, 2.0, and 0.6 US cents for \texttt{gpt-5}, \texttt{gpt-5-mini}, and \texttt{gpt-5-nano}. A more detailed breakdown stratified by retrieval difficulty is in Appendix~\ref{app:ee}.

\section{Discussion}

Multi-modal RAG and OCR-free visual retrieval aim at the same goal: grounding generation on document evidence rather than model memory. Both decouple knowledge from weights and benefit from fine-grained, late-interaction matching. Where they differ is \emph{when} visual content is interpreted. Modality-converting pipelines interpret earlier (via OCR/layout analysis and summarization). Visual retrievers interpret later (the LLM reads retrieved page images). This placement matters in practice.

\textit{Experiment 1 (capacity $\times$ pipeline).}
When model capacity is limited, direct visual retrieval underperforms. In our initial screen, \texttt{gemma-3-27b-it} did poorly with ColPali compared with text or multi-modal conversion. This shows a failure mode: the system can retrieve the right page but the generator cannot reliably read it. With a stronger model (GPT-4o), the gap closes and ColPali becomes competitive. The design lesson is conservative: conversion to text lowers the burden on the generator; OCR-free pipelines lean on LLM visual reasoning.

\textit{Experiment 2 (retriever $\times$ model family).}
Within the GPT-5 family, retriever choice mattered less than model scale. ColPali, ColQwen, and ColFlor delivered near-identical means with overlapping CIs. ColFlor matched ColPali while being far smaller and faster, making it an efficient default with frontier models. Using ColFlor, \texttt{gpt-5} exceeded \texttt{gpt-5-mini} by only about 1–2\% on average, while \texttt{gpt-5-mini} is roughly \emph{5×} cheaper per token. For many deployments, that trade-off favors \texttt{gpt-5-mini} with ColFlor.

\textit{Pipeline simplicity vs.\ robustness.}
Visual retrievers avoid OCR and fragile PDF parsing (table reconstruction, layout heuristics), simplifying ingestion and limiting parser-induced errors. The cost is a higher reliance on the LLM’s visual reasoning. Conversion-based pipelines add engineering steps and may lose detail, but generalize better to mid-sized models and yield clearer evidence traces.

\textit{Implication.}
Choose capacity-aware designs. Under ample compute or hosted APIs, OCR-free retrieval is attractive. Under tighter budgets or on-prem constraints, conversion-based multi-modal RAG remains the safer and more interpretable default.

\subsection{Limitations}
This study has several limitations. The benchmark is small (120 MCQs) and focused on glycobiology, so broader validation in other biomedical subfields (e.g., radiology, genomics) is needed. The item mix is skewed toward “easy” cases (57.5\%), which reduces power for medium and hard analyses; we treat those strata as exploratory. We checked for benchmark contamination answer-order permutation, however we plan on a more robust strategy in the future, such as paraphrasing questions, adding non-relevant retrieval information, generating non correct answers. We analyzed one open-source model in depth (\texttt{gemma-3-27b-it}) because its initial performance was superior to other options; future work should span a wider range of open-source VLMs to map scaling and architecture effects. Finally, we reported multiple-choice accuracy; adding factual consistency and evidence-use metrics would give a fuller behavioral picture.

\subsection{Future Research Directions}

\textit{Retriever design.}
We will systematically compare modern visual retrievers—ColPali\cite{faysse2025colpali}, ColQwen\cite{manu_colqwen2_v0_2}, and ColFlor\cite{masry2024colflor}—and newer vision/video variants to see how retrieval detail (page-, patch-, or token-level) and index compression affect accuracy, speed, and storage on visually rich biomedical articles. Prior work shows that late-interaction, OCR-free retrievers can be both simple and highly competitive, which makes them strong candidates for evidence retrieval in practice. We will evaluate on document-centric and vision-centric benchmarks to ensure results transfer beyond a single dataset. 

\textit{Lightweight and cost-aware MM-RAG.}
We will pair smaller, faster retrievers with mid-sized generators to reduce cost and latency while tracking accuracy and “cost per correct answer.” ColFlor is an example of a compact, OCR-free retriever (about 174M parameters) that approaches the performance of heavier models like ColPali, but runs substantially faster, suggesting good cost–utility for institutional deployments.

\textit{Trustworthy grounding.}
Beyond accuracy, we will stress-test how reliably models use retrieved evidence—especially images, tables, and complex page layouts. Recent vision-centric studies find that even strong models under-use retrieved visuals, highlighting the need for strict citation, showing the exact snippets or page crops used, and simple checks that link each answer back to its sources. These measures support auditability in clinical and research settings. 

\textit{Grounded platform for researchers.}
We intend to turn these ideas into a simple, secure platform that helps researchers in their own domain ask better questions, find the right evidence, and draft stronger proposals. The system will use RAG so every answer is \emph{grounded} in trusted sources the user selects/provides (papers, guidelines, lab documents, PDFs—including figures and tables). For visuals, it will either convert images/tables to text or pass the page image—whichever is clearer—and always show the exact snippets used so claims can be checked. The platform will scale from a single lab to an institution, support on-premise options to protect sensitive data, and let teams choose smaller or larger models to match budget and accuracy needs. 

\section{Conclusion}
This study compared text-centric and OCR-free visual retrieval strategies for multi-modal grounding on a glycobiology MCQ benchmark. We found a clear capacity-dependent pattern: conversion-based pipelines (text and multi-modal summaries) are more reliable with mid-size VLMs (e.g., Gemma-3-27B-IT), whereas OCR-free late-interaction retrieval becomes competitive with frontier models (e.g., GPT-4o / GPT-5). Among visual-retrievers, ColFlor matched ColPali under GPT-5 while offering a smaller footprint, suggesting an attractive efficiency-accuracy trade-off. Practically, these results argue for capacity-aware MM-RAG design: convert earlier when model visual reasoning is constrained; retrieve page images directly when model capacity allows.

Our work is preliminary. The benchmark is small and single-domain. Broader evaluations across modalities and domain will test the generality of these findings and support trustworthy multi-modal assistants in biomedicine. 

\section*{GENAI USAGE DISCLOSURE}
In this manuscript we used Gemini and ChatGPT to help improve its language and overall presentation. Our aim with these tools was to enhance the text's clarity, coherence, and correctness, and we carefully reviewed and edited any suggestions they provided. We want to clearly state that all scientific content, core ideas, and analyses are entirely our own original work, developed without relying on Gemini or ChatGPT for any conceptual or analytical aspects of the research. As authors, we are responsible for the manuscript's final content and affirm the originality of its scientific contributions stemming from our own efforts.

\section*{Code Availability}
A reproducibility package (benchmark template, scripts and prompts) are available at: \href{https://github.com/pkocbek/multi-modal-RAG-biomed}{https://github.com/pkocbek/multi-modal-RAG-biomed}.
\bibliographystyle{IEEEtran}
\bibliography{references_cleaned}

\appendices
\section{List of doi links for the 25 original research and review manuscripts} \label{app:dd}
\begin{scriptsize}
\begin{verbatim}
https://doi.org/10.1186/s12967-018-1695-0
https://doi.org/10.1097/hjh.0000000000002963
https://doi.org/10.1186/s12967-018-1616-2
https://doi.org/10.3390%2Fbiom13020375
https://doi.org/10.1016/j.bbagen.2017.06.020
https://doi.org/10.1172%2Fjci.insight.89703
https://doi.org/10.1016%2Fj.isci.2022.103897
https://doi.org/10.1002%2Fart.39273
https://doi.org/10.1016/j.bbadis.2018.03.018
https://doi.org/10.1097/MIB.0000000000000372
https://doi.org/10.1053%2Fj.gastro.2018.01.002
https://doi.org/10.1186/s13075-017-1389-7
https://doi.org/10.1021/pr400589m
https://doi.org/10.1161/CIRCRESAHA.117.312174
https://doi.org/10.2337/dc22-0833
https://doi.org/10.1097%2FMD.0000000000003379
https://doi.org/10.1158/1078-0432.CCR-15-1867
https://doi.org/10.1093/gerona/glt190
https://doi.org/10.1111/imr.13407
https://doi.org/10.1053/j.gastro.2018.05.030
https://doi.org/10.1016/j.csbj.2024.03.008
https://doi.org/10.1016/j.cellimm.2018.07.009
https://doi.org/10.1016/j.biotechadv.2023.108169
https://doi.org/10.4049/jimmunol.2400447
\end{verbatim}
\end{scriptsize}

\section{General prompt template was used for creating table and figure summaries} \label{app:aa}
\begin{small}
\begin{verbatim}
You are an AI assistant specialized 
in summarizing tables and figures for
efficient retrieval. \n\nInstructions:
\n\nIdentify Input Type: Explicitly 
state whether the input provided is a
table or a figure.\nScientific Abstract:
Summarize the contents concisely in the 
style of a scientific abstract. Include
relevant numeric values and key findings. 
\nRetrieval Optimization: Structure your 
summary clearly, optimizing keywords and 
phrasing to enhance retrieval and indexing.
\nLength Constraint: Your summary must
strictly adhere to a maximum of 300 words 
or 250 tokens. Do not exceed this limit 
under any circumstances. Any text 
exceeding will be just cutoff post 
generation.\nAvoid Generic Openings: Do 
not start your summary with generic 
phrases such as "The image provided is,
" "The table shows," or similar 
introductory sentences. Instead, 
immediately describe the core content.
\nPrevent Redundancy: Write succinctly, 
avoiding repetition of concepts or data 
points.\nFinal output: Only summary text.
If no relevant data is present, 
output \'\'.
\end{verbatim}
\end{small}
\section{Example docker configurations for Docling, Qdrant and gemma-3-27b-it} \label{app:bb}
Dockling:
\begin{scriptsize}
\begin{verbatim}
docker run 	\
    --rm  --gpus all \
    -e DOCLING_SERVE_ENABLE_UI=true \
    -e DOCLING_SERVE_MAX_SYNC_WAIT=600 \
    -e DOCLING_SERVE_ENABLE_REMOTE_SERVICES=true \
    -e "DOCLING_SERVE_API_KEY=${DOCLING_API_KEY}" \
    --name docling_serve \
    -p 5001:5001 \
    ghcr.io/docling-project/docling-serve-cu124:latest
\end{verbatim}
\end{scriptsize}

Qdrant:
\begin{scriptsize}
\begin{verbatim}
docker run \
	--rm \
    --name qdrant_vd \
	--gpus=all \
	-p 6333:6333 \
	-p 6334:6334 \
	-e QDRANT__GPU__INDEXING=1 \
    -e "QDRANT__SERVICE__API_KEY=${QDRANT_API_KEY}" \
    --ulimit nofile=65536:65536 \
    -v ./src/vectordb/storage:/qdrant/storage \
    -v ./src/vectordb/custom_config.yaml:
    /qdrant/config/custom_config.yaml \
	qdrant/qdrant:gpu-nvidia-latest
\end{verbatim}
\end{scriptsize}

Gemma-3-27B-IT (vLLM):
\begin{scriptsize}
\begin{verbatim}
docker run --gpus all -it --rm --pull=always \
    --name gemma_27b \
    -v "${HF_DIR}:/root/.cache/huggingface" \
    --env "HUGGING_FACE_HUB_TOKEN=
    ${HUGGING_FACE_HUB_TOKEN}" \
    --env TRANSFORMERS_OFFLINE=1 \
    --env VLLM_RPC_TIMEOUT=180000 \
    --env HF_DATASET_OFFLINE=1 \
    -p 8006:8000 \
    --ipc=host \
    vllm/vllm-openai:latest \
    --model "google/gemma-3-27b-it" \
    --limit_mm_per_prompt '{"image": 8}' \
    --gpu-memory-utilization 0.82 \\
    --max_model_len 16000 \
    --enable-sleep-mode 
\end{verbatim}
\end{scriptsize}

\section{General prompt template for evaluation} \label{app:cc}
\begin{small}
\begin{verbatim}
Generate a JSON with the query_answer, 
the answer provided behind the letters: 
A, B, C, and D. These are the values. 
Additional information if provided in
the Context below. If the Context is 
not empty, analyse it and choose from 
the letters. MAKE SURE your output is 
one of the four values stated. 
Here is the query: {question}. 
Here are the choices: {question_string} 
    Context:

\end{verbatim}
\end{small}
\section{Performance of visual retrievers across the GPT-5 model family stratified by retrieval difficulty. }\label{app:ee}

\begin{table}[htbp]
\centering
\caption{Precision@5 across the GPT-5 model family with bootstrapped 95\% CIs.}
\label{tab:p5_appendix1}
\renewcommand{\arraystretch}{1.1}
\setlength\tabcolsep{1.2 pt}
\begin{tabular}{llrrrr}
\hline
\begin{tabular}{c}\textbf{LLM}\\ \textbf{model}\end{tabular} &
\begin{tabular}{c}\textbf{Retrieval}\\ \textbf{model}\end{tabular} &
\begin{tabular}{c}\textbf{Easy}\\ (n=69)\end{tabular} &
\begin{tabular}{c}\textbf{Medium}\\ (n=24)\end{tabular} &
\begin{tabular}{c}\textbf{Hard}\\ (n=27)\end{tabular} &
\begin{tabular}{c}\textbf{Average}\\ (n=120)\end{tabular} \\
\hline

\multirow{3}{*}{\textbf{gpt-5}}
& ColPali & 0.022 & 0.017 & 0.018 & 0.020 \\
&         & \scriptsize{[0.000, 0.115]} & \scriptsize{[0.000, 0.070]} &
            \scriptsize{[0.000, 0.049]} & \scriptsize{[0.000, 0.094]} \\
& ColQwen & 0.032 & 0.003 & 0.027 & 0.025 \\
&         & \scriptsize{[0.023, 0.042]} & \scriptsize{[0.000, 0.015]} &
            \scriptsize{[0.019, 0.034]} & \scriptsize{[0.017, 0.033]} \\
& ColFlor & 0.021 & 0.017 & 0.010 & 0.018 \\
&         & \scriptsize{[0.021, 0.021]} & \scriptsize{[0.017, 0.017]} &
            \scriptsize{[0.010, 0.010]} & \scriptsize{[0.018, 0.018]} \\
\hline

\multirow{3}{*}{\begin{tabular}{c}\textbf{gpt-5}\\ \textbf{-mini}\end{tabular}}
& ColPali & 0.021 & 0.025 & 0.020 & 0.022 \\
&         & \scriptsize{[0.011, 0.032]} & \scriptsize{[0.000, 0.056]} &
            \scriptsize{[0.000, 0.045]} & \scriptsize{[0.017, 0.026]} \\
& ColQwen & 0.031 & 0.004 & 0.025 & 0.024 \\
&         & \scriptsize{[0.026, 0.037]} & \scriptsize{[0.000, 0.015]} &
            \scriptsize{[0.025, 0.025]} & \scriptsize{[0.019, 0.030]} \\
& ColFlor & 0.021 & 0.017 & 0.010 & 0.018 \\
&         & \scriptsize{[0.021, 0.021]} & \scriptsize{[0.017, 0.017]} &
            \scriptsize{[0.010, 0.010]} & \scriptsize{[0.018, 0.018]} \\
\hline

\multirow{3}{*}{\begin{tabular}{c}\textbf{gpt-5}\\ \textbf{-nano}\end{tabular}}
& ColPali & 0.027 & 0.017 & 0.012 & 0.022 \\
&         & \scriptsize{[0.014, 0.041]} & \scriptsize{[0.017, 0.017]} &
            \scriptsize{[0.000, 0.031]} & \scriptsize{[0.011, 0.034]} \\
& ColQwen & 0.033 & 0.006 & 0.027 & 0.026 \\
&         & \scriptsize{[0.027, 0.038]} & \scriptsize{[0.000, 0.012]} &
            \scriptsize{[0.019, 0.034]} & \scriptsize{[0.024, 0.028]} \\
& ColFlor & 0.021 & 0.017 & 0.010 & 0.018 \\
&         & \scriptsize{[0.021, 0.021]} & \scriptsize{[0.017, 0.017]} &
            \scriptsize{[0.010, 0.010]} & \scriptsize{[0.018, 0.018]} \\
\hline

\end{tabular}
\end{table}

\begin{table}[htbp]
\centering
\caption{Latency per query (s) across the GPT-5 model family with bootstrapped 95\% CIs.}
\label{tab:elapsed_appendix}
\renewcommand{\arraystretch}{1.1}
\setlength\tabcolsep{1.2 pt}
\begin{tabular}{llrrrr}
\hline
\begin{tabular}{c}\textbf{LLM}\\ \textbf{model}\end{tabular} &
\begin{tabular}{c}\textbf{Retrieval}\\ \textbf{model}\end{tabular} &
\begin{tabular}{c}\textbf{Easy}\\ (n=69)\end{tabular} &
\begin{tabular}{c}\textbf{Medium}\\ (n=24)\end{tabular} &
\begin{tabular}{c}\textbf{Hard}\\ (n=27)\end{tabular} &
\begin{tabular}{c}\textbf{Average}\\ (n=120)\end{tabular} \\
\hline

\multirow{3}{*}{\textbf{gpt-5}}
& ColPali & 20.50 & 19.31 & 19.93 & 20.12 \\
&         & \scriptsize{[0.00, 153.60]} & \scriptsize{[0.00, 175.57]} &
            \scriptsize{[0.00, 179.20]} & \scriptsize{[0.00, 163.66]} \\
& ColQwen & 22.89 & 22.15 & 23.11 & 22.77 \\
&         & \scriptsize{[0.00, 81.84]} & \scriptsize{[0.00, 78.91]} &
            \scriptsize{[0.00, 83.48]} & \scriptsize{[0.00, 81.50]} \\
& ColFlor & 18.02 & 18.63 & 16.37 & 17.83 \\
&         & \scriptsize{[0.00, 154.33]} & \scriptsize{[0.00, 154.54]} &
            \scriptsize{[0.00, 141.62]} & \scriptsize{[0.00, 151.84]} \\
\hline

\multirow{3}{*}{\begin{tabular}{c}\textbf{gpt-5}\\ \textbf{-mini}\end{tabular}}
& ColPali & 13.85 & 13.26 & 14.26 & 13.80 \\
&         & \scriptsize{[0.00, 44.10]} & \scriptsize{[0.00, 41.85]} &
            \scriptsize{[0.00, 48.99]} & \scriptsize{[0.00, 44.56]} \\
& ColQwen & 24.26 & 24.46 & 24.45 & 24.34 \\
&         & \scriptsize{[0.00, 88.52]} & \scriptsize{[0.00, 91.19]} &
            \scriptsize{[0.00, 90.60]} & \scriptsize{[0.00, 89.54]} \\
& ColFlor & 21.72 & 21.81 & 19.68 & 21.33 \\
&         & \scriptsize{[0.00, 89.65]} & \scriptsize{[0.00, 83.66]} &
            \scriptsize{[0.00, 75.96]} & \scriptsize{[0.00, 85.56]} \\
\hline

\multirow{3}{*}{\begin{tabular}{c}\textbf{gpt-5}\\ \textbf{-nano}\end{tabular}}
& ColPali & 14.12 & 14.39 & 13.64 & 14.09 \\
&         & \scriptsize{[0.00, 48.75]} & \scriptsize{[0.00, 50.35]} &
            \scriptsize{[0.00, 43.99]} & \scriptsize{[0.00, 48.16]} \\
& ColQwen & 23.44 & 24.30 & 23.57 & 23.66 \\
&         & \scriptsize{[0.00, 77.03]} & \scriptsize{[0.00, 81.46]} &
            \scriptsize{[0.00, 79.67]} & \scriptsize{[0.00, 78.55]} \\
& ColFlor & 13.10 & 13.54 & 12.45 & 13.07 \\
&         & \scriptsize{[0.00, 44.53]} & \scriptsize{[0.00, 45.47]} &
            \scriptsize{[0.00, 43.28]} & \scriptsize{[0.00, 44.49]} \\
\hline

\end{tabular}
\end{table}

\begin{table}[htbp]
\centering
\caption{Tokens per query across the GPT-5 model family with bootstrapped 95\% CIs.}
\label{tab:tokens_appendix}
\renewcommand{\arraystretch}{1.2}
\setlength\tabcolsep{1.2 pt}
\begin{tabular}{llrrrr}
\hline
\begin{tabular}{c}\textbf{LLM}\\ \textbf{model}\end{tabular} &
\begin{tabular}{c}\textbf{Retrieval}\\ \textbf{model}\end{tabular} &
\begin{tabular}{c}\textbf{Easy}\\ (n=69)\end{tabular} &
\begin{tabular}{c}\textbf{Medium}\\ (n=24)\end{tabular} &
\begin{tabular}{c}\textbf{Hard}\\ (n=27)\end{tabular} &
\begin{tabular}{c}\textbf{Average}\\ (n=120)\end{tabular} \\
\hline

\multirow{3}{*}{\textbf{gpt-5}}
& ColPali & 4588.51 & 4431.44 & 5037.90 & 4643.05 \\
&         & \tiny{[4171.88, 5005.15]} & \tiny{[3524.13, 5338.76]} &
            \tiny{[4686.62, 5389.17]} & \tiny{[4269.59, 5016.51]} \\
& ColQwen & 4296.22 & 4392.33 & 4633.54 & 4385.31 \\
&         & \tiny{[4177.71, 4414.73]} & \tiny{[4249.02, 4535.65]} &
            \tiny{[4407.49, 4859.59]} & \tiny{[4273.68, 4496.94]} \\
& ColFlor & 4570.55 & 4544.65 & 4878.21 & 4626.25 \\
&         & \tiny{[4181.45, 4959.66]} & \tiny{[4333.58, 4755.71]} &
            \tiny{[4522.44, 5233.98]} & \tiny{[4378.85, 4873.66]} \\
\hline

\multirow{3}{*}{\begin{tabular}{c}\textbf{gpt-5}\\ \textbf{-mini}\end{tabular}}
& ColPali & 7764.92 & 7766.67 & 7874.39 & 7787.21 \\
&         & \tiny{[7212.10, 8317.74]} & \tiny{[7101.18, 8432.15]} &
            \tiny{[6572.19, 9176.59]} & \tiny{[7112.34, 8462.08]} \\
& ColQwen & 8827.38 & 8908.42 & 9202.65 & 8920.67 \\
&         & \tiny{[8681.47, 8973.29]} & \tiny{[8821.23, 8995.61]} &
            \tiny{[9031.50, 9373.81]} & \tiny{[8867.94, 8973.40]} \\
& ColFlor & 7604.41 & 7910.82 & 7202.44 & 7592.96 \\
&         & \tiny{[7498.25, 7710.57]} & \tiny{[7880.70, 7940.93]} &
            \tiny{[6778.63, 7626.24]} & \tiny{[7562.46, 7623.45]} \\
\hline

\multirow{3}{*}{\begin{tabular}{c}\textbf{gpt-5}\\ \textbf{-nano}\end{tabular}}
& ColPali & 9634.73 & 10141.63 & 10439.17 & 9909.67 \\
&         & \tiny{[9303.77, 9965.69]} & \tiny{[9735.92, 10547.34]} &
            \tiny{[8831.33, 12047.01]} & \tiny{[9408.38, 10410.96]} \\
& ColQwen & 11601.02 & 11705.74 & 12255.69 & 11755.52 \\
&         & \tiny{[11469.15, 11732.89]} & \tiny{[11689.46, 11722.03]} &
            \tiny{[12200.02, 12311.37]} & \tiny{[11688.70, 11822.33]} \\
& ColFlor & 9927.24 & 10378.53 & 9941.10 & 10031.55 \\
&         & \tiny{[9916.64, 9937.85]} & \tiny{[10103.47, 10653.59]} &
            \tiny{[9650.87, 10231.32]} & \tiny{[9928.94, 10134.17]} \\
\hline

\end{tabular}
\end{table}

\begin{table}[htbp]
\centering
\caption{Throughput per query across the GPT-5 model family with bootstrapped 95\% CIs.}
\label{tab:latency_appendix}
\renewcommand{\arraystretch}{1.2}
\setlength\tabcolsep{1.2 pt}
\begin{tabular}{llrrrr}
\hline
\begin{tabular}{c}\textbf{LLM}\\ \textbf{model}\end{tabular} &
\begin{tabular}{c}\textbf{Retrieval}\\ \textbf{model}\end{tabular} &
\begin{tabular}{c}\textbf{Easy}\\ (n=69)\end{tabular} &
\begin{tabular}{c}\textbf{Medium}\\ (n=24)\end{tabular} &
\begin{tabular}{c}\textbf{Hard}\\ (n=27)\end{tabular} &
\begin{tabular}{c}\textbf{Average}\\ (n=120)\end{tabular} \\
\hline

\multirow{3}{*}{\textbf{gpt-5}}
& ColPali & 405.45 & 415.76 & 455.99 & 417.88 \\
&         & \scriptsize{[0.00, 3561.15]} & \scriptsize{[0.00, 3761.30]} &
            \scriptsize{[0.00, 4225.25]} & \scriptsize{[0.00, 3739.00]} \\
& ColQwen & 353.52 & 383.96 & 382.56 & 366.17 \\
&         & \scriptsize{[0.00, 931.45]} & \scriptsize{[0.00, 1005.56]} &
            \scriptsize{[0.00, 1000.34]} & \scriptsize{[0.00, 960.92]} \\
& ColFlor & 420.42 & 390.53 & 507.63 & 431.14 \\
&         & \scriptsize{[0.00, 3678.67]} & \scriptsize{[0.00, 3355.48]} &
            \scriptsize{[0.00, 4422.60]} & \scriptsize{[0.00, 3754.74]} \\
\hline

\multirow{3}{*}{\begin{tabular}{c}\textbf{gpt-5}\\ \textbf{-mini}\end{tabular}}
& ColPali & 939.25 & 984.49 & 1004.85 & 962.55 \\
&         & \scriptsize{[0.00, 2336.21]} & \scriptsize{[0.00, 2395.39]} &
            \scriptsize{[0.00, 2577.26]} & \scriptsize{[0.00, 2396.77]} \\
& ColQwen & 726.47 & 794.62 & 789.15 & 754.34 \\
&         & \scriptsize{[0.00, 1952.89]} & \scriptsize{[0.00, 2228.31]} &
            \scriptsize{[0.00, 2144.00]} & \scriptsize{[0.00, 2050.25]} \\
& ColFlor & 379.56 & 391.09 & 409.73 & 388.19 \\
&         & \scriptsize{[0.00, 1495.23]} & \scriptsize{[0.00, 1403.71]} &
            \scriptsize{[0.00, 1239.65]} & \scriptsize{[0.00, 1423.52]} \\
\hline

\multirow{3}{*}{\begin{tabular}{c}\textbf{gpt-5}\\ \textbf{-nano}\end{tabular}}
& ColPali & 1282.09 & 1292.56 & 1278.30 & 1283.69 \\
&         & \scriptsize{[0.00, 3431.41]} & \scriptsize{[0.00, 3398.66]} &
            \scriptsize{[0.00, 3280.06]} & \scriptsize{[0.00, 3391.69]} \\
& ColQwen & 849.58 & 847.58 & 938.78 & 866.97 \\
&         & \scriptsize{[0.00, 2304.27]} & \scriptsize{[0.00, 2327.62]} &
            \scriptsize{[0.00, 2580.72]} & \scriptsize{[0.00, 2364.71]} \\
& ColFlor & 1332.50 & 1324.49 & 1495.30 & 1363.26 \\
&         & \scriptsize{[0.00, 3426.45]} & \scriptsize{[0.00, 3408.69]} &
            \scriptsize{[0.00, 3940.36]} & \scriptsize{[0.00, 3523.50]} \\
\hline

\end{tabular}
\end{table}

\begin{table}[htbp]
\centering
\caption{Cost (USD) per run across the GPT-5 model family with with bootstrapped 95\% CIs.}
\label{tab:price_appendix}
\renewcommand{\arraystretch}{1.2}
\setlength\tabcolsep{1.2 pt}
\begin{tabular}{llrrrr}
\hline
\begin{tabular}{c}\textbf{LLM}\\ \textbf{model}\end{tabular} &
\begin{tabular}{c}\textbf{Retrieval}\\ \textbf{model}\end{tabular} &
\begin{tabular}{c}\textbf{Easy}\\ (n=69)\end{tabular} &
\begin{tabular}{c}\textbf{Medium}\\ (n=24)\end{tabular} &
\begin{tabular}{c}\textbf{Hard}\\ (n=27)\end{tabular} &
\begin{tabular}{c}\textbf{Average}\\ (n=120)\end{tabular} \\
\hline

\multirow{3}{*}{\textbf{gpt-5}}
& ColPali & 3.17 & 1.20 & 1.21 & 5.57 \\
&         & \scriptsize{[2.88, 3.45]} & \scriptsize{[0.95, 1.44]} &
            \scriptsize{[1.13, 1.29]} & \scriptsize{[5.12, 6.02]} \\
& ColQwen & 2.96 & 1.19 & 1.11 & 5.26 \\
&         & \scriptsize{[2.88, 3.05]} & \scriptsize{[1.15, 1.22]} &
            \scriptsize{[1.06, 1.17]} & \scriptsize{[5.13, 5.40]} \\
& ColFlor & 3.15 & 1.23 & 1.17 & 5.55 \\
&         & \scriptsize{[2.89, 3.42]} & \scriptsize{[1.17, 1.28]} &
            \scriptsize{[1.09, 1.26]} & \scriptsize{[5.26, 5.85]} \\
\hline

\multirow{3}{*}{\begin{tabular}{c}\textbf{gpt-5}\\ \textbf{-mini}\end{tabular}}
& ColPali & 1.07 & 0.42 & 0.38 & 1.87 \\
&         & \scriptsize{[1.00, 1.15]} & \scriptsize{[0.38, 0.46]} &
            \scriptsize{[0.32, 0.44]} & \scriptsize{[1.71, 2.03]} \\
& ColQwen & 1.22 & 0.48 & 0.44 & 2.14 \\
&         & \scriptsize{[1.20, 1.24]} & \scriptsize{[0.48, 0.49]} &
            \scriptsize{[0.43, 0.45]} & \scriptsize{[2.13, 2.15]} \\
& ColFlor & 1.05 & 0.43 & 0.35 & 1.82 \\
&         & \scriptsize{[1.04, 1.06]} & \scriptsize{[0.43, 0.43]} &
            \scriptsize{[0.33, 0.37]} & \scriptsize{[1.82, 1.83]} \\
\hline

\multirow{3}{*}{\begin{tabular}{c}\textbf{gpt-5}\\ \textbf{-nano}\end{tabular}}
& ColPali & 0.27 & 0.11 & 0.10 & 0.48 \\
&         & \scriptsize{[0.26, 0.28]} & \scriptsize{[0.11, 0.11]} &
            \scriptsize{[0.09, 0.12]} & \scriptsize{[0.45, 0.50]} \\
& ColQwen & 0.32 & 0.13 & 0.12 & 0.56 \\
&         & \scriptsize{[0.32, 0.32]} & \scriptsize{[0.13, 0.13]} &
            \scriptsize{[0.12, 0.12]} & \scriptsize{[0.56, 0.57]} \\
& ColFlor & 0.27 & 0.11 & 0.10 & 0.48 \\
&         & \scriptsize{[0.27, 0.27]} & \scriptsize{[0.11, 0.12]} &
            \scriptsize{[0.09, 0.10]} & \scriptsize{[0.48, 0.49]} \\
\hline

\end{tabular}
\end{table}

\begin{table}[htbp]
\centering
\caption{Price-per-correct answer (US cents) across the GPT-5 model family with bootstrapped 95\% CIs.}
\label{tab:price_correct_appendix}
\renewcommand{\arraystretch}{1.2}
\setlength\tabcolsep{1.2 pt}
\begin{tabular}{llrrrr}
\hline
\begin{tabular}{c}\textbf{LLM}\\ \textbf{model}\end{tabular} &
\begin{tabular}{c}\textbf{Retrieval}\\ \textbf{model}\end{tabular} &
\begin{tabular}{c}\textbf{Easy}\\ (n=69)\end{tabular} &
\begin{tabular}{c}\textbf{Medium}\\ (n=24)\end{tabular} &
\begin{tabular}{c}\textbf{Hard}\\ (n=27)\end{tabular} &
\begin{tabular}{c}\textbf{Average}\\ (n=120)\end{tabular} \\
\hline

\multirow{3}{*}{\textbf{gpt-5}}
& ColPali & 5.61 & 5.70 & 6.05 & 5.72 \\
&         & \scriptsize{[3.21, 8.01]} & \scriptsize{[4.53, 6.86]} &
            \scriptsize{[5.62, 6.47]} & \scriptsize{[4.14, 7.29]} \\
& ColQwen & 5.08 & 5.31 & 5.85 & 5.28 \\
&         & \scriptsize{[4.82, 5.34]} & \scriptsize{[4.95, 5.67]} &
            \scriptsize{[5.57, 6.14]} & \scriptsize{[5.08, 5.48]} \\
& ColFlor & 5.74 & 5.46 & 5.72 & 5.67 \\
&         & \scriptsize{[4.90, 6.57]} & \scriptsize{[4.17, 6.74]} &
            \scriptsize{[3.53, 7.90]} & \scriptsize{[5.36, 5.97]} \\
\hline

\multirow{3}{*}{\begin{tabular}{c}\textbf{gpt-5}\\ \textbf{-mini}\end{tabular}}
& ColPali & 1.84 & 1.88 & 1.97 & 1.87 \\
&         & \scriptsize{[1.64, 2.04]} & \scriptsize{[1.61, 2.15]} &
            \scriptsize{[1.29, 2.65]} & \scriptsize{[1.59, 2.15]} \\
& ColQwen & 2.10 & 2.13 & 2.84 & 2.23 \\
&         & \scriptsize{[1.92, 2.29]} & \scriptsize{[1.76, 2.50]} &
            \scriptsize{[2.11, 3.56]} & \scriptsize{[1.98, 2.47]} \\
& ColFlor & 1.88 & 2.03 & 1.82 & 1.90 \\
&         & \scriptsize{[1.42, 2.33]} & \scriptsize{[2.03, 2.04]} &
            \scriptsize{[1.71, 1.93]} & \scriptsize{[1.66, 2.14]} \\
\hline

\multirow{3}{*}{\begin{tabular}{c}\textbf{gpt-5}\\ \textbf{-nano}\end{tabular}}
& ColPali & 0.50 & 0.54 & 0.69 & 0.54 \\
&         & \scriptsize{[0.42, 0.58]} & \scriptsize{[0.49, 0.59]} &
            \scriptsize{[0.61, 0.76]} & \scriptsize{[0.48, 0.60]} \\
& ColQwen & 0.61 & 0.67 & 0.91 & 0.67 \\
&         & \scriptsize{[0.57, 0.65]} & \scriptsize{[0.58, 0.75]} &
            \scriptsize{[0.74, 1.08]} & \scriptsize{[0.62, 0.72]} \\
& ColFlor & 0.51 & 0.54 & 0.71 & 0.55 \\
&         & \scriptsize{[0.45, 0.57]} & \scriptsize{[0.49, 0.60]} &
            \scriptsize{[0.45, 0.96]} & \scriptsize{[0.48, 0.62]} \\
\hline

\end{tabular}
\end{table}

\end{document}